# Sensor-based, Time-critical Mobility of Autonomous Robots in Cluttered Spaces: A Harmonic Potential Approach

Ahmad A. Masoud, Ali Al-Shaikhi,

Electrical Engineering, King Fahad University of Petroleum & Minerals, Dhahran, Saudi Arabia, masoud@kfupm.edu.sa, shaikhi@kfupm.edu.sa,

Abstract: This paper suggests an integrated navigation system for an unmanned ground vehicle operating in an unknown cluttered environment. The navigator supports time-critical mobility making it possible for a mobile robot to reach a target from the first attempt without the need for a dedicated exploration and mapping stage. The robot only uses necessary and sufficient egocentric local sensory data collected on its way to the target. The construction of the navigation structure revolves around key properties of the harmonic potential field approach to motion planning. The robot's trajectory is well-behaved and direct-to-the-goal. It contains only the minimum number of detours necessary to accommodate the sensory data and maintain the robot in a safe, goal-oriented state. The navigation structure is developed and its theoretical basis is explained. Extensive experimental validation of its properties and performance is carried-out using the X80 robotic platform.

List of symbols

| | |
|---|---|
| HPF: | Harmonic potential field |
| TCM: | Time critical mobility |
| G-Type | Geno-type |
| P-Type | Pheno-Type |
| $V(X)$ | Harmonic potential |
| $X$, $\dot{X}(X)$ | Agent actual position and velocity vectors |
| $\dot{X}_r(X)$ | Agent reference velocity |
| $\lambda$ | Orientation vector of agent |
| $X_T$ | Target position |
| $E(X)$ | Guidance field |
| $\gamma_{X0}(X)$ | Hypothesis plan |
| $\Pi$ | Operation perimeter |
| $\Omega$ | Actual Usable space |
| $\overline{\Omega}$ | Believed to be Usable known space |
| $\Psi$ | Common components between actual and belief usable space ($\Psi = \Omega \cap \overline{\Omega}$) |
| $O$ | Hazardous regions |
| $\overline{O}$ | Believed to be hazardous regions |
| $O_s$ | Safety events detected by the sensors at a certain instant in time |
| $O_h$ | Event retested in the on-demand, safety-based map ($O_h \supseteq O_s$) |
| $U(\ )$ | Control |
| $\Xi r$ | Region robot occupies at time t |
| $\Xi s$ | Region sensed by robot at time t |
| $P(\ )$ | Functional that measures the energy a guidance policy consumes |
| $\nabla$ | Gradient operator |
| $\nabla \cdot$ | Divergence operator |
| $\theta$ | agent orientation angle in 2D space |
| $vc$ | Tangential control speed of the robot in 2D space |
| $\omega c$ | Angular control speed of the robot in 2D space |
| $v$ | Tangential velocity of the robot in 2D space |
| $\omega$ | Angular speed of the robot in 2D space |
| $\omega_R$, $\omega_L$ | Right and left wheel speeds of a differential drive robot |
| $\omega_h$, $\phi$ | Driving wheel speed and steering angle of a car-like robot |
| $\Delta\theta$ | Angle between actual robot velocity and desired robot velocity in 2D space |
| $\eta_d, \eta_c$ | Cosine and sine $\Delta\theta$ respectively |
| $S(t)$ | Distance measurement signal from ultrasonic sensor |
| $DSE(i,j)$ | Discrete representation array covering $\Pi$ (i and j are the discrete space indices) |
| $\Delta$ | Spatial resolution of $DSE(i,j)$ |

# I. Introduction

There is a growing interest in developing robotics aids to help first responders in firsthand evaluation of incidents [1, 2, 3]. The information obtained by such systems is critical and serves as the primary link in a chain of information exchange that leads to making important decisions. To be useful, these aids have to provide the human operator with situational awareness in a timely and constrained manner. There are other strict and challenging properties this type of robots needs to satisfy [4]. In a first responder situation, a mobile robot should be able to move to a designated area in an unstructured and unknown environment. Robot deployment cannot always rely on accurate maps since they are often not available. The robot should be able to function under zero *a priori* knowledge without engaging in time-consuming activities reserved for exploration and mapping only. The entire robot's effort should be dedicated to reaching the zone of interest using the necessary and sufficient information its sensors pick-up while it is heading to the target. The robot should also be able to accommodate any available *a priori* information in its database. This information is used to accelerate convergence and enhance performance. Other requirements complicate the deployment of such robots. For example, a robot of this type must have an agile and robust behavior that is communication-aware. First responders are always risk averse. They tolerate no technology that increases the level of mission uncertainty. Therefore, constraints on the behavior of the robot, which include a guarantee that it can reach the target in a relatively intuitive and predictable manner, are necessary. A first responder environment is usually hazardous. The probability of damage, even loss, is high. Therefore, a first responder robot should not be too expensive. The above requirements represent a challenge not only to the procedures of using these devices, but also to the existing paradigms for autonomous mobility generation themselves.

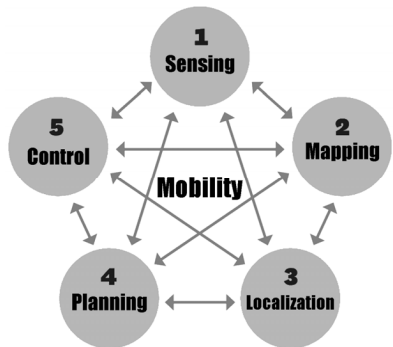

Figure-1: Networked components of a mobility structure

Mobility is a composite activity that emerges from the networking of basis activity modules (Figure-1). One of these modules is the sensing module, which is concerned with the acquisition of environment data [5]. The representation module creates a map by processing and structuring this data [6, 7]. A localization module [8] establishes a correspondence between the information available at the agent's current location on the map with the actual information that exists at its location in the environment. The guidance module provides the agent with the direction along which it has to proceed (a know what to do signal) in order to reach the target [9,10]. The control module [11, 12] converts the reference direction into a control signal that is fed to the agent's actuators (a know how to do signal). The norm seems to study a linear stack of these modules (Figure-2) assuming that their integration in one mobility system is possible by individually optimizing the performance of each component.

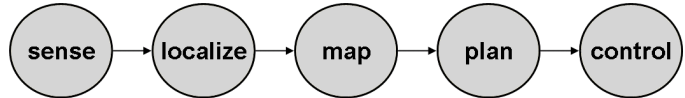

Figure-2: Linearly stacked mobility modules

The assumption that system-level reliability is achievable by making the components individually reliable is questionable, to say the least. It is a well-known fact that reliable systems are constructible from unreliable components [13, 14] and vice versa. There is a growing concern that a stacked, modularized view of mobility can at best lead to an overly complicated system with modest performance. A mobility system is more than its components. It also consists of a host structure to network its parts properly. More importantly, the components must be inter-conditioned to enable the modules to operate as a collective within the confines of the host structure. This leads to an overly complex joint design problem. Many guidelines exist in the literature for constructing mobility structures [15-18]. Despite the intensive work on the integration of mobility modules, many issues relating to how they function are open areas of research.

Algorithmic, sensor-based guidance procedures (e.g. the bug algorithm [19-20] ) are usually prime candidates when information about the environment is intermittent and scarce. They are provably-correct, model-free and provide directly-usable information in the local reference frame of the robot. However, many obstacles prevent their use in a practical situation. For example, the bug algorithm requires a robot to move tangentially to an obstacle's surface. Most realistic obstacles have rough surfaces and their tangent may not exist. While the Bug procedure can provide provably-correct guidance performance in two dimensional spaces, there are no guarantees that the kinodynamic path the controller generates is provably-correct. Most importantly, the trajectories are far from being practical, let alone close to optimal.

The alternative to algorithmic sensor-based methods is model-based techniques. To the best of the authors' knowledge existing model-based techniques, adopt the linearly stacked scheme in connecting the mobility modules (figure-2). In particular, they seem to dissociate map-building activities from goal-oriented mobility. In time critical mobility (TCM), dedicating time exclusively to exploration and map building is not admissible. These methods attempt to objectify the navigation process in two ways. First, they assume a world coordinate system that a robot has to relate its situation to. They also seem to consider an accurate and full spatial characterization of the environment as a major factor in the ability of an agent to project a successful mobility action. In a first responder situation, it is not possible to accurately and (or) fully spatially characterize the components of an environment. It is also difficult to transform reliably data back and forth between the local coordinates of the robot and the global coordinates of the representation.

The trend of developing theoretical frameworks that depart from the linear stack approach and jointly examine the construction of more than one mobility module is growing. Examples of this are: simultaneous localization and mapping [21], direct guidance from sensory (observation) space [22], joint guidance and control [23]. To the best of the authors' knowledge, a theoretical framework that jointly tackles all the modules needed for providing an autonomous agent with mobility does not exist. Putting together a complete mobility system seems to be mainly dependant on the experience of the designer [24,25].

While a formal procedure for designing mobility systems seems currently out of reach, this work focus on the role the planning module [26-32] plays in order to provide an answer on how to tackle TCM. Although a planner is only a component of a mobility system, it is the provider of the reference motion policy (Figure-3) an agent needs to realize in order to function. If the mobility system is working properly, the behavior of the agent will closely follow that reference. To support the construction of a mobility system for a TCM, a planner must also have the ability to accommodate the manner in which the other mobility modules function. In other words, there exist means of sensing, localization, mapping and control which when used in the mobility system, the agent projects the desired behavior provided by the planner. As shown in this paper, harmonic potential field-based (HPFs) planners [33,34] are capable of playing this role.

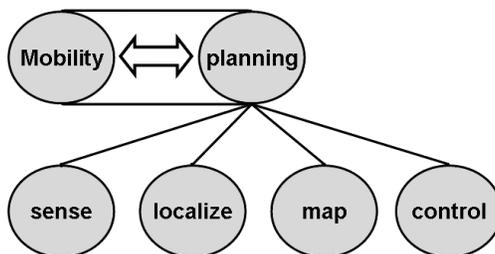

Figure-3: Planning as a reference for mobility

An HPF can efficiently host the sensing, representation, guidance and control modules to cerate an integrated, servo-level navigation action that suits time-critical missions. Experiments clearly demonstrate a promising ability of the suggested HPF-based mobility system to utilize successfully, virtually unprocessed, local, egocentric data even from difficult-to-use sensors such as ultrasonic sensors [35]. Harmonic potential field-based planning techniques are efficient, versatile and provably-correct means for inducing context-sensitive, goal-oriented, intelligent and constrained behavior in an autonomous agent. Despite their deep mathematical structure [36] and strong relation to connectionist artificial intelligence (AI) and hybrid AI [37-39], they are easily understood using simple physical processes such as fluid flow [40-42], electric current movement [43,44] and mechanical stress propagation in solids [45]. A basic setting of the HPF approach was suggested by Connolly et. al [46] who solved a Laplace boundary value problem in the Dirichlet setting. There are different boundary conditions one may use to generate the guidance field [47]. Each one of these settings has distinct topological properties [48]. Boundary conditions function to factor-in the context into the behavior of the agent. HPF-based planners are better understood in the context of hybrid, partial differential equation – ordinary differential equation systems (PDE-ODE) [49]. The PDE part of the system functions to create a dense family of potential guidance plans (figure-4) for any situation the agent may be in (guidance policy). On the other hand, the ODE component of the system has the much-limited role of selecting and executing one of these plans. Harmonic functions have many useful properties for motion planning. Most notably, a harmonic potential is also a Morse function [50,51] and a general form of the potential field-based navigation function suggested in [52]. The HPF approach can operate in a model-based and (or) sensor-based mode. It can also accommodate a variety of motion constraints. Hybrid PDE-ODE systems operate in conformity with the artificial life (AL) approach to behavior synthesis [53]. The process of morphogenesis [54] is responsible for inducing a mission-compliant guidance structure on the substrate of the vector guidance elements that is covering situation space (figure-5).

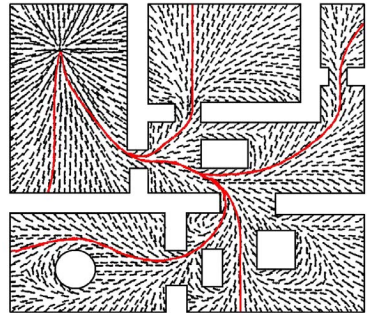
Figure-4: Harmonic potential field-based guidance policy

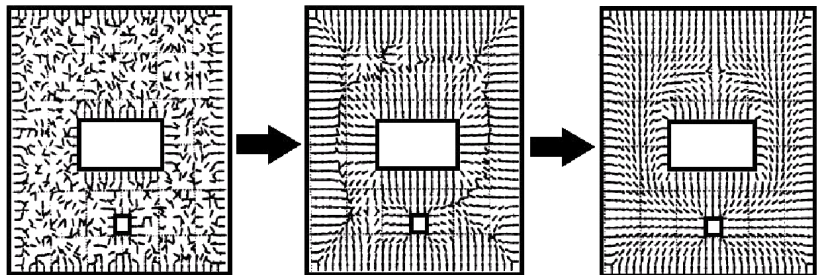
Figure-5: Evolution of an HPF-based guidance policy

The HPF approach is an excellent paradigm for constructing motion-planning techniques. For example, the HPF-based technique in [55] makes it possible to jointly incorporate directional constraints along with regional avoidance constraints in a provably-correct manner to plan a path to a target point. The variant of the HPF in [56] provides a provably-correct procedure for taking into consideration inherent ambiguities that prevent the partitioning of an environment into admissible and forbidden regions. The HPF approach has the ability to plan motion on discrete graphs [57]. It can also generate planning procedures that operate in a communication-aware mode [58]. Experimental assessment of the approach reveals that it is well-suited for guiding autonomous agents in challenging real-life terrains [59]. It demonstrates excellent ability to interface with other components of an integrated navigation system [60]. It is also extendable, in a provably-correct manner, to the case of active intelligent targets [61]. There is a wide spectrum of techniques for computing a harmonic potential. Some of the major numerical approaches for generating an HPF are the finite difference methods (FDM) [62], the finite element method (FEM) [63] and the boundary element method (BEM). BEM is the fastest and can generate a continuous, 2D harmonic potential in real-time (few milliseconds) [64]. Hardware computational means of HPF do exist, for example, HPF-based neural nets [65,66]. In addition, FPGA-based computational means [67] can generate an HPF on a 50x50 grid in less than 100 microseconds. Analog VLSI's [68-71], which in case of memristors [72,73], have the ability to generate an HPF in a fraction of a microsecond.

This paper describes the construction of a reliable and inexpensive servo-level mobility system that is built around the properties of the HPF planning paradigm and suits time-sensitive mobility. A safety-based representation of the environment [74] is used that intertwines sensory data collection with goal-oriented motion. The system is able to steer a mobile robot to a specified target along an obstacle-free path in a fully unknown environment. It functions to keep the robot in a safe, goal-oriented state and bypass the need to engage in a phase that is exclusively dedicated to exploration and mapping. The work demonstrates that time critical-mobility, which suits a first responder situation, is possible from the first attempt using impoverished egocentric sensing. An affordable robotics platform, the X80, is used to extensively test the suggested mobility system. The test experiments use low-end sensing and localization. On-line data is collected using only one ultrasonic sensor (the front sensor of the X80). Naïve dead-reckoning is used for localization where the pose of the robot is computed by directly integrating its linear and angular speeds. All processing is carried-out on-board a host computer. Sensory and control signals are exchanged in real-time between the host and the X80 platform using a wireless communication link.

Problem setting is contained in section II. Section III discusses the belief-based, time-sensitive mobility paradigm. Section IV presents the representation, guidance and control modules used in constructing the TCM mobility system along with the procedure used to integrate them into one system. Section V provides a comprehensive set of sensor-based experiments carried out on the X80 UGV platform to demonstrate the capabilities of the approach. Conclusions are placed in section VI.

## II. Problem Setting

In the suggested mobility scenario an emergency crew arriving at a certain location would only provide a mobile robot with the perimeter of operation ($\Pi$) and the target location in this perimeter both specified with respect to robots axis at its initial location. The mobility system is expected to guarantee that the robot will safely reach the designated location in a timely and constrained manner. The type of unmanned ground vehicles (UGVs) considered is described by the equation (1):

$$\begin{bmatrix} \dot{x} \\ \dot{y} \\ \dot{\theta} \end{bmatrix} = H(\begin{bmatrix} x \\ y \\ \theta \end{bmatrix})\begin{bmatrix} v \\ \omega \end{bmatrix}, \quad \begin{bmatrix} v \\ \omega \end{bmatrix} = G(U) \tag{1}$$

where x and y are the center of the robot in the local coordinates, $\theta$ is the orientation of the robot, v and $\omega$ are the tangential and angular speeds of the robot respectively and U is the control signal vector. The model in equation-1 accommodates a wide class of

UGVs including the two important types (Figure-6): the differential drive robot (2) and the car-like robot (3) described by the system equations:

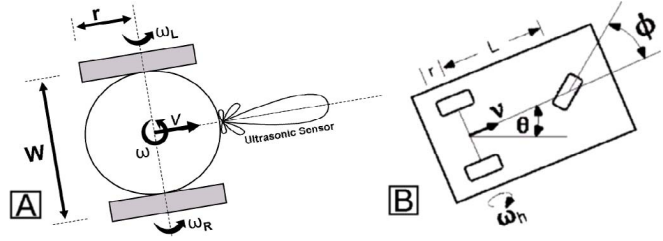

Figure-6: Differential drive (A) & Car-like (B) robots

$$\begin{bmatrix} \dot{x} \\ \dot{y} \\ \dot{\theta} \end{bmatrix} = \begin{bmatrix} \cos(\theta) & 0 \\ \sin(\theta) & 0 \\ 0 & 1 \end{bmatrix} \begin{bmatrix} v \\ \omega \end{bmatrix}, \quad \begin{bmatrix} v \\ \omega \end{bmatrix} = \begin{bmatrix} \dfrac{r}{2} & \dfrac{r}{2} \\ \dfrac{r}{W} & \dfrac{-r}{W} \end{bmatrix} \begin{bmatrix} \omega_R \\ \omega_L \end{bmatrix}, \quad (2)$$

$$\begin{bmatrix} \dot{x} \\ \dot{y} \\ \dot{\theta} \end{bmatrix} = \begin{bmatrix} \cos(\theta) & 0 \\ \sin(\theta) & 0 \\ 0 & 1 \end{bmatrix} \begin{bmatrix} v \\ \omega \end{bmatrix}, \quad \begin{bmatrix} v \\ \omega \end{bmatrix} = \begin{bmatrix} r \cdot \omega_h \\ \tan(\phi) \dfrac{r \cdot \omega_h}{L} \end{bmatrix}, \quad (3)$$

where $\omega_L$ and $\omega_R$ are the right and left wheel speeds of the differential drive robot, W is its width and r is the wheels' radius, $\omega_h$ is the speed of the driving wheel of the car-like robot, $\varphi$ is the steering angle and L is the distance between the center of the driving wheels axis and the steering wheel.

The TCM system is required to synthesize the mobility policy (4):

$$U(X, X_T, \overline{\Omega}(t)) \quad X \in \Pi \quad (4)$$

Such that
$$\lim_{t \to \infty} X(t) \to X_T,$$
$$\Omega \supset \Xi r(t) \quad \forall t$$

and
$$\|\gamma_{Xo}\| < K \cdot \|X(0) - X_T\|$$

where $\Xi r(t)$ is the region occupied by the robot at time t, $\Omega$ is the actual unknown free space contained in the perimeter, $\overline{\Omega}$ is the free space registered in the map of the robot at time t, K is a finite positive number that is expected not to be much larger than unity and $\gamma_{Xo}$ is the trajectory (hypothesis plan) of the agent from an initial starting point Xo to the target $X_T$ (5),

$$\gamma_{X0} = \{X : \frac{d\gamma_{X0}}{dt} = \dot{X}(X(t)), \quad \gamma_{X0}(X(t_0)) = X0\}. \quad (5)$$

## III. On-demand, Belief-based, Time-Sensitive Mobility

While serious challenges face objective model-based techniques when dealing with a first responder situation, subjective belief-based methods (figure-7) have a reasonable ability to handle the difficulties caused by the scarcity of information such a situation impose. This approach treats a representation (a subjective map) as a self-referential belief that evolves under the influence of the sensory data collected during mission execution. What guides the robot's actions is a hypothesis plan that is a member of a dense field of plans (guidance policy, figure-4) that is constructed given the belief the agent has about its situation. By design, the guidance policy guarantees that at any location in the space which the agent believe is capable of supporting motion, there exists a hypothesis plan to take the robot to the target. If at a certain point in space the sensory feedback falsifies the hypothesis plan, the agent switches to another hypothesis plan that will move it from its current location in space to the target. A belief-based mobility system need only maintain continuity of action towards the goal and ensure that the agent is always in a state where the components of the environment do not inhibit its ability to function. Any environment component that does not adversely affect the ability of the robot to act is considered as a do not care component whose presence, let alone geometry, need not be factored in robot mobility. Enforcing the above two requirements enables an agent to carryout its mission even if the environment in which it is operating is initially unknown.

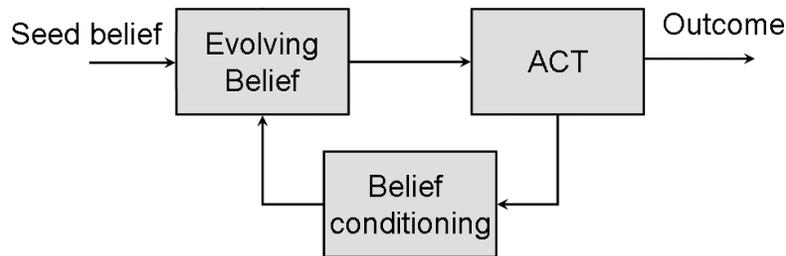

Figure-7: Belief-based causality

Dedicating too much effort and resources to accurate characterization of the spatial confines of the environment components (e.g. hazards, obstacles etc.) is most probably the cause of many mobility artifacts. In spatial navigation, task-centered safety, not whole domain geometry, is the enabler of mobility. In the absence of information, space should be optimistically looked-at as a resource usable in its entirety for supporting mobility. Mapping has to proceed in a safety-centered and on-demand manner relative to the interim motion policy governing the agent's behavior. The belief-based approach to mobility acquires and process only the subset of environmental information needed to test the validity of the hypothesis plan of action adopted by the robot. It is desirable that this set be necessary and sufficient to carry out the task in an acceptable manner. Necessary and sufficient information is concentrated around the hypothetical trajectory the robot attempts to move along to the target. Intertwining goal-oriented actuation with sensing (figure-8) significantly reduces the requirement from that of registering all components of space to registering only the local components surrounding the one-dimensional trajectory of the robot (one-dimensional sensing). This subset of space that is cooperatively generated because of interaction between actuation and sensing contains the necessary and sufficient information to carryout the job, given the agent's initial motion policy. As can be seen from figure-8, the subset of information collected by the agent on its way to the target is usually sparse. It could even be the empty set if the initial hypothesis plan that is constructed using zero knowledge of the components of the environment happens to lay a safe path to the target. It ought to be mentioned that the above deviates from the current mindset in navigation that considers three-dimensional (or at least two-dimensional) sensing essential for attaining navigation competence [85].

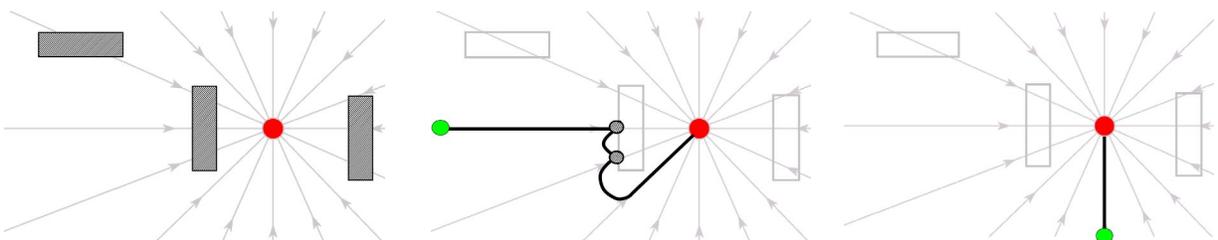

Figure-8: Necessary and sufficient sensing relative to a mobility policy

Considering a plan false depends on the online sensory feedback degrading confidence in the safety of the location that the plan will lead the robot towards. If that happens, the potentially unsafe location is marked as a blocked region in the representation. The blocking causes reduction in the confidence of the region's local neighborhood to support motion. Consequently, the updated motion policy discourages motion in these regions by steering the agent as far as possible from the blocked zones.

The subset of space that the guidance policy intends to direct the robot through is assessed using the local, egocentric sensor field of the robot. Timely assessment of that region requires either relaxing the accuracy of the estimates or utilizing specialized fast hardware to produce high precision estimates in a timely manner. Since hardware affordability is an issue, the latter is not an option. Making the robot block any zone that generates a reasonable level of suspicion of being unsafe is a simple and effective way of achieving this objective. It is faster and more reliable than attempting to determine accurately the boundary of the unsafe regions. While conservatively using the sensory data to assess the safety of a location may cause the loss of potentially usable safe space, it highly enhances the robot's safety and significantly reduces the amount of computations. It also enables the utilization of inexpensive local sensing (e.g. ultrasonic sensing) in generating practical robot mobility. Successful mobility only requires that the sensor interpretation process does not falsely place the agent under the belief that it is surrounded by hazard and isolated from the target. In other words, if $\overline{\Omega}$ is the safe space that the robot believes it can move through and $\Omega$ is the actual space that the robot can move through, these spaces need only to satisfy the conditions (6):

$$\Psi = \Omega \cap \overline{\Omega}$$
$$X_T \in \Psi$$
(6)

and $\Psi$ is a connected region.

This is by no means a restrictive assumption since one may initially select $\overline{\Omega}$ as the whole perimeter set ($\Pi$) in which the robot may operate. Experiments show that losing the ability to reach the target due to the accumulation of false positive hazardous space is unlikely to happen even if the robot initially has no information about its content. In the unlikely event that this happens, simple countermeasures, e.g. resetting $\overline{\Omega}$ to $\Pi$, are enough to rectify the problem.

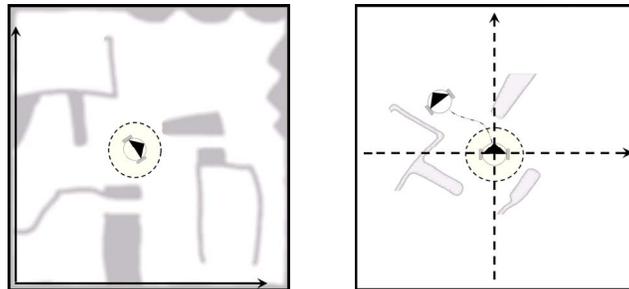

Figure-9: World coordinates versus subjective, robot-centered coordinates

Another issue related to subjectivity is the construction of a reference coordinate system for logging the sensory data and indexing the components of the representation. In a first responder situation, the environment is unlikely to support the construction of a reasonably reliable global, reference coordinate system. The environment is also expected to be GPS-denied and (or) RF-challenged, preventing localization through radio navigation. The only feasible option in this case is for the robot to operate in a self-referential coordinate system (figure-9). The location of the agent from which it starts to explore a certain region in space is selected as the origin of the subjective coordinate system. The principal axis of the robot may be selected as one axis of the coordinate system (the positive x-axis) while the other axis is selected to form a right-hand orthogonal coordinate system. The subjective, self-referential frame of reference makes it possible to synthesize action and log sensory data directly in the local coordinates of the robot bypassing the need to do any transformation.

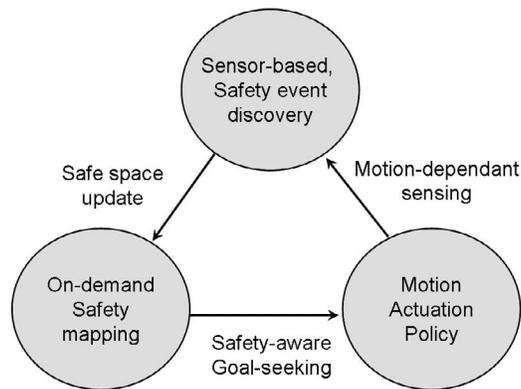

Figure-10: On-demand, mobility-dependant, safety-based representation

It is clear that the suggested mapping procedure and goal-oriented mobility are interrelated. The initial policy of action, that is based on the agent's initial belief of what the environment contains, determines how much sensing and computation are needed to get the job done. This interrelation gives rise to the concept of on-demand, safety-based mapping (figure-10). As seen in figure-8, if by chance the agent starts motion from a position where the trajectory assigned by the actuation policy is feasible, the sensors will record no safety-related events. Hence, there is no need for data processing and action adjustment. The agent will reach the target with its initial belief and actuation policy unadjusted. The interconnection between sensing and belief-based goal-oriented mobility can guarantee that a potentially reachable target is attainable. It also uses the necessary and sufficient information for doing so given the initial belief of the agent. This in turn means that the adjustments to the actuation policy are necessary and sufficient for executing the task.

To support the above, the actuation module must generate a whole-domain actuation policy providing a control command for every possible point covered by the representation. This policy has to be optimal, in some sense, given the amount of information that is initially available to the agent. For example, when no information is available, the agent assumes that the perimeter ($\Pi$) is fully usable for navigation. The actuation policy will drive the robot to the target along a straight line. The disruption to the actuation policy due to the accommodation of newly discovered environment components has to be kept to the minimum needed to preserve agent's safety and its ability to continue seeking the goal. The ability to operate in this manner enables one to make the statement that the effort the robot projects to reach the target is optimal, given the initial information. Requiring minimum disruption to the actuation policy by the sensory data is a key factor in solving important problems facing the sensor-based operation of a mobile robot. It adds a strong element of predictability to mobility in unknown spaces, an issue that is crucial in a first responder situation.

Restricting the data intake of the mobility system to that provided by impoverished sensing does result in a system level problem. The cause of this problem is the incompatibility of the nature of the sensors with the nature of the actuators. The output of an impoverished sensor is always noisy and intermittent (undefined on random time intervals). On the other hand, the actuation signal should be well behaved and available all the time (figure-11). Treating a representation as an evolving, whole-domain belief that is converted into whole-domain guidance by the HPF approach creates a strong buffer between sensing and actuation and bypasses this problem.

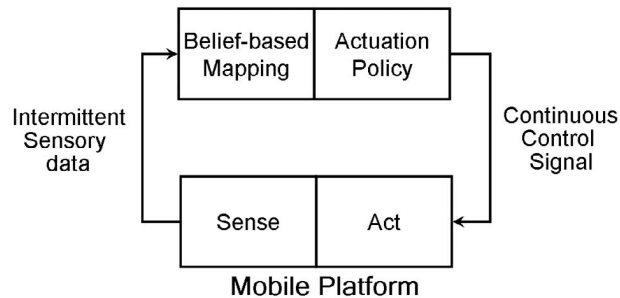
Figure-11: Conflicting nature of sensing and actuation

## IV Mobility system components

This section presents the mapping, guidance and control components used to build the TCM system. The presentation emphasizes the collective and distributed nature in which the agent maintains a safe, goal-oriented state.

A. On-demand, safety-based mapping:
On-demand mapping makes use of the fact that if an agent is always in a goal-oriented state, not all the content of the environment is relevant for executing the mission. The only components that matter are those that inhibit the execution of the hypothesis plans the agent selects from the motion policy. The motion policy need only guarantee the agent's safety and continuity of action given the belief of what the environment contents are. Sensory feedback continuously and locally assesses the validity of the selected hypothesis plan. The sensor cover must contain the region that the agent is going to occupy in the immediate future (figure-12). Whenever the validity of a hypothesis plan is in question, the offending environment component is mapped into the belief-based safety map. The map is then processed to create a motion policy that would transfer the agent to another hypothesis plan whose deficiency is still unconfirmed. The successful testing of a hypothesis plan in a manner that allows the agent to recover from a possible safety event depends on the ability to sense the region where the plan will drive the robot towards in the immediate future. Reasonably accurate determination of this region is not difficult if the agent knows its current state and the action that it is going to assume in the near future. The environment components within the sensory zone that are identified as inhibitors of mobility are considered necessary information that is added to the evolving representation. Along with focusing sensory activities on the region that the robot is going to occupy in the near future, the system conservatively registers the hazardous components of the environment in that region if they exist. The following explains in more details the construction of the on-demand, safety-centered representation.

let $\Xi_r(X(t))$ be the region which the robot occupies at time t and $\Xi_s(X(t))$ be the region covered by the sensors of the robot at time t. Let O be the region at which the contents of the environment inhibit the operation of the robot and $O_s(t)$ the unsafe regions detected by the sensors at time t. Also, let $\overline{O}$ be the subset of space the representation marks as unsafe. Notice that this set can be the empty set if the robot has no initial information about the environment. It is assumed that a reasonably accurate estimate of the immediate future space which the agent is going to occupy ($\Xi_r(X(t+\Delta T))$) is possible using the mobility policy of the agent and its state. It is necessary and sufficient that the sensors of the robot cover that zone (7)

$$\Xi_s(X(t)) \supseteq \Xi_r(X(t+\Delta T)) \tag{7}$$

The safety representation is constructed as follows (8): let

$$O_s(t) = O \cap \Xi_s(X(t)) \tag{8}$$

Depending on the level of confidence needed to exclude hazardous region, construct an enlarged region ($O_h(t)$) that contains $O_s(t)$ (9)

$$O_h(t) \supseteq O_s(t) \tag{9}$$

Finally, add the enlarged region to the safety representation (10)

$$\overline{O}(t+\Delta T) = \overline{O}(t) \cup O_h(t). \tag{10}$$

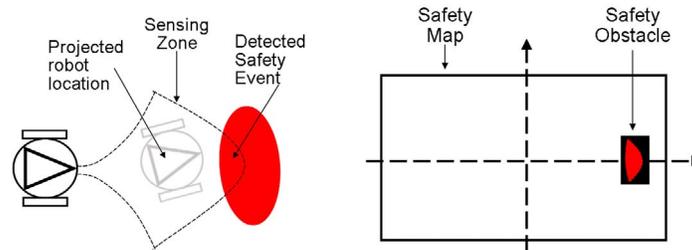

Figure-12: On-demand, safety-based sensing

B. The HPF-based Guidance Module:

The harmonic potential field planner in the Dirichlet setting is selected for generating the goal-seeking guidance action for the mobility system. The generating BVP is (11):

solve: $\quad\quad\quad\quad\quad\quad\quad\quad \nabla^2 V(X) \equiv 0 \quad\quad X \in \Omega \tag{11}$
subject to: $\quad\quad\quad\quad\quad V(X) = 1$ at $X = O$ and $V(X_T) = 0$,
$\quad\quad\quad\quad\quad\quad\quad\quad \dot{X} = -\nabla V(X) \quad\quad X(0) \in \Omega$

where V is the harmonic potential, $\nabla^2$ is the Laplace operator, $\nabla$ is the gradient operator, O is the hazardous space, $\Omega$ is the usable space ($\Omega = \Pi - O$), $\Pi$ is the operation perimeter and $X_T$ is the target point.

Connolly showed that a harmonic potential provides a probabilistic measure of collision [75]. Setting V(X)=1 at the hazardous regions is equivalent to indicating to the system that collision at that location in space is expected with probability=1. The Dirichlet setting accommodates this information, in a safety-aware manner, to construct the guidance policy (figure-13). The policy instructs the agent to increase strictly its distance away from the regions where collision is eminent (i.e. move strictly along the normal path away from the obstacle). The guidance policy enhances the safety of the agent by continuously steering it to the target along directions that lead to maximum decrease in the collision probability. From a safety point of view, the approach yields results comparable to those of a Vornoi-based approach [76].

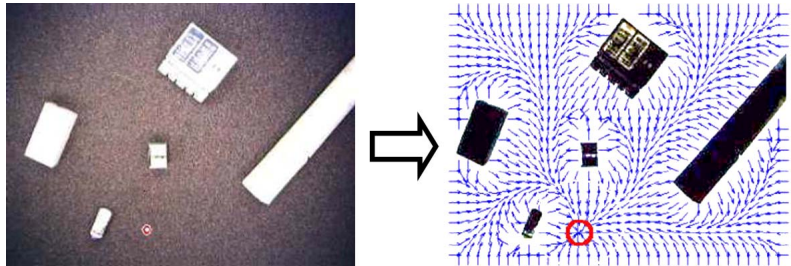
Figure-13: Safety-sensitive guidance – HPF Dirichlet setting

Online computation of the guidance policy is a challenging issue. This is due to the requirement that the policy fully covers the representation grid. The dimensions of a practical representation grid used by an autonomous agent are usually in hundreds even thousands of pixels. The information in the representation grid has to be processed in real-time by the agent's modest, on-board computational means. This makes online re-computation of the guidance policy difficult each time the representation is adjusted. Harmonic potential-based guidance policies have a noncommittal nature and do not suffer from this problem. A change in an HPF-based guidance policy due to local updates does not globally propagate through space. Rather, it remains effectively confined to the local vicinity of the change. Figure-14 shows an HPF-based navigation policy and the same policy updated with a point obstacle. It also shows a plot measuring the difference between the guidance fields in both cases. As can be seen, the change in the guidance policy rapidly drops away from the updated region. It is straightforward to prove this property mathematically. According to the harmonic superposition principle, the difference between the two harmonic functions before and after the update is a harmonic function. This harmonic function is also subject to Dirichlet boundary conditions with zero value at the old boundaries and a positive constant at the point of local update. It is obvious that the surface surrounding the updated region is much smaller than the surface of the previously existing regions. A simple application of the divergence theorem shows that the difference potential will rapidly drop as one moves away from the area of the change.

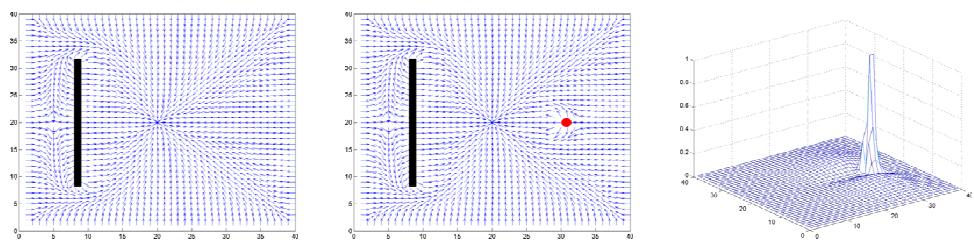
Figure-14: HPF localizes disturbance of in guidance policies

The consequence of the above is that a local representation update will only yield a local guidance policy update. In other words, the computational burden needed to generate the updated policy under the influence of the sensory data is decoupled from the size of the representation grid. Instead, it is coupled to the size and frequency of the updates. As mentioned earlier, several efficient means do exist for whole domain computation of the HPF-based guidance field. The ability to generate the modified guidance policy through spatially localized computation has a pronounced effect on the ability to reduce system complexity and enhance its performance.

Another consequence of the quick fading of update-induced disturbance in the mobility policy is that the disruptive effect of noisy sensing on motion is marginalized (Figure-15).

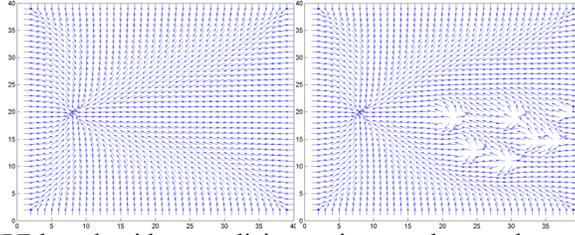
Figure-15: HPF-based guidance policies retain smoothness when sensing is noisy

An HPF-based guidance produces a least action motion policy. This directly follows from the Dirichlet principle [77] where equation (11) is a minimizer of the energy functional (12)

$$P(X) = \int_\Omega |\nabla V(X)|^2 dX \qquad X \in \Omega \qquad (12)$$

This in turn means that an HPF-based guidance policy can be adjusted with minimum disruption to the modified guidance field. Moreover, one can make the statement that the behavior generated is optimal, given the amount of information initially provided to the robot about its environment.

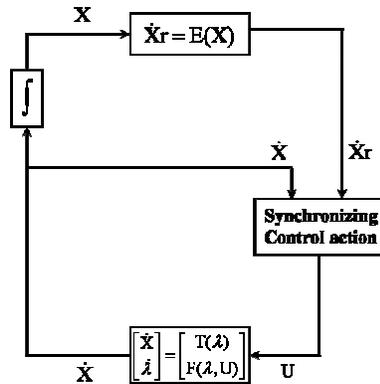
Figure-16: Virtual velocity attractor approach for converting an HPF-based guidance element into a control signal

C. Control policy from guidance policy

The guidance stage provides, at any given location in space, the corresponding direction along which the robot must move if it is to reach its target. An agent has to actualize this reference direction by deploying its actuators. The system equation that governs the relation between motion and actuation for a large class of mobile agents is (13):

$$\begin{bmatrix} \dot{X} \\ \dot{\lambda} \end{bmatrix} = \begin{bmatrix} T(\lambda) \\ F(\lambda, U) \end{bmatrix} \qquad (13)$$

where X is the location vector of the agent, $\lambda$ is its orientation vector and U is a vector that contains the control variables. Making guidance and control coalesce to produce the desired mobility pattern is an excessively complex, joint design problem. The method in [21] suggests an approach to manage, in a provably-correct manner, the complexity of generating a control policy that is in aim with an HPF-based guidance policy. The approach indirectly controls the spatial variables by controlling their time derivatives. It uses the HPF-based guidance policy as a reference velocity field ($\dot{X}_r$) for the agent. At each position in space, the control acts to align the velocity of the agent ($\dot{X}$) with the corresponding reference velocity (Figure-16). If the actual velocity of the agent and the reference velocity from the guidance field are initially in phase, the kinodynamic trajectory, which evolves under the influence of the control, is identical to the kinematic trajectory marked by the guidance field. In effect, the combination of the guidance policy and the synchronization control is equivalent to an on-demand control policy that generates the actuation signal when and where it is needed (figure-17).

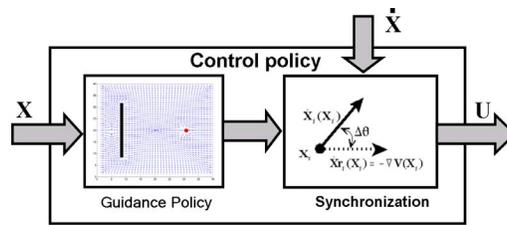

Figure-17: On-demand control policies from HPF-based guidance polices and a virtual velocity synchronizer

The above procedure tackles static guidance policies where the environment is *a priori* known and the policy does not change during the robot's operation. In a TCM situation, the guidance policy keeps changing under the influence of the sensors. As a result, the guarantees that the synchronization method can prevent collision are no longer valid. One way of using the method in a provably-correct manner is to require the robot to stop each time the guidance policy gets updated. The robot then re-aligns itself along the new reference velocity. Frequent stopping is at odds with the requirement that the robot exhibits time-sensitive mobility. It also leads to excessive energy consumption [78, 79], increases unpredictability of the robot's behavior and places considerable stress on its hardware.

Provably-correct adjustment of the virtual velocity synchronization approach to deal with online sensory updates is possible. It makes use of a safety-based interpretation of the relative orientation of the robot's velocity with respect to the reference velocity from the guidance policy. If the velocity of the robot is in phase with the reference velocity, the robot is in a low risk state (figure-18). On the other hand, if the velocities are in an antipodal configuration, the robot is in a high-risk state. It is important to notice that only the tangential component of the robot's speed has significant influence on safety. The effect of the angular component is marginal, even negligible (if the robot has a cylindrical frame). Based on the above, it is possible to reduce the probability of collision, even eliminate it, if the synchronizing controller properly modulates the tangential velocity of the robot. The modulation process sets the tangential velocity at maximum when the robot is perfectly aligned with the reference velocity. The velocity is set at a minimum, preferably zero, when the robot and the reference velocity are in an antipodal configuration.

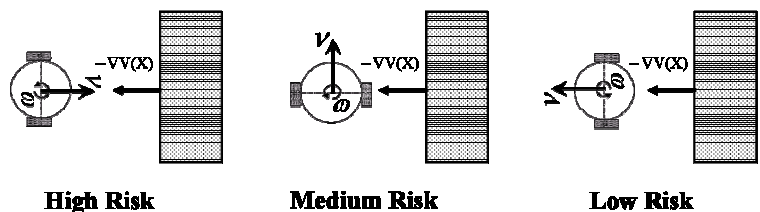

Figure-18: Safety-aware control generating

The development of the safety-aware, on-demand navigation control policy uses a version of the velocity synchronization approach designed for use with UGVs [80]. The following presents a practical variant of the method in [80] for aligning the robot's velocity with the desired one in the presence of sensory update. The procedure for generating the mobility control signal involves three issues. The issues are safety, hardware friendliness and computation in the local coordinates of the robot.

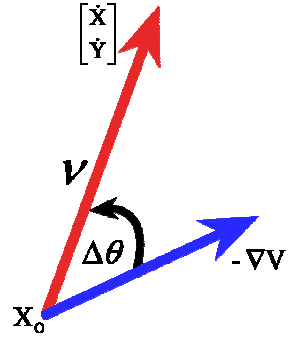

Figure-19: Controller aligns guidance signal with robot's velocity

First the sine ($\eta_c$) and cosine ($\eta_d$) of the angle between (Figure-19) the velocity of the robot and the guidance vector ($\Delta\theta$) along with the straight line distance to the target (dst) are computed (14):

$$\eta_d = \cos(\Delta\theta) = \frac{Ex \cdot \dot{x} + Ey \cdot \dot{y}}{\sqrt{\dot{x}^2 + \dot{y}^2}},$$

$$\eta_c = \sin(\Delta\theta) = \frac{Ex \cdot \dot{y} - Ey \cdot \dot{x}}{\sqrt{\dot{x}^2 + \dot{y}^2}}$$

$$dst = \sqrt{(x - x_T)^2 + (y - y_T)^2}$$

(14)

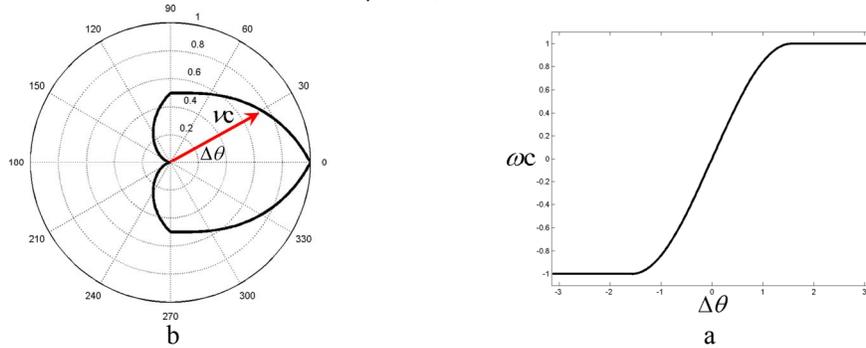

b            a

Figure-20: Desired tangential and angular control velocities of the robot respectively

A control angular actuation signal ($\omega c$) must function to synchronize the actual robot speed with the desired guidance vector. A form of the angular synchronizing signal (Figure-20a) is (15):

$$\omega c = \omega d \cdot \begin{bmatrix} \eta_c & \eta_d > 0 \\ +1 & \eta_c > 0 \,\&\, \eta_d < 0 \\ -1 & \eta_c < 0 \,\&\, \eta_d < 0 \end{bmatrix},$$

(15)

where $\omega d$ is the maximum desired angular speed the robot should assume. A control tangential robot speed ($vc$) should be maximum when the robot is in phase with the guidance vector. This speed should gradually drop as the speed gets out of phase with guidance. $vc$ should drop to zero when the robot's speed and guidance are in an antipodal configuration which is created by the detection of an event that endangers the safety of the robot (Figure-20b). The robot should start to slow-down when it is at an Rc distance from the target. Its value should be zero at the target location. The speed may have the form (16):

$$vc = vd \cdot \begin{bmatrix} (\eta_d + 1)/2 & \eta_d < 0 \\ 1 - (|\omega c / \omega d|)/2 & \eta_d \geq 0 \\ dst/Rc & dst \leq Rc \end{bmatrix} \quad (16)$$

where vd is the maximum tangential speed the robot should assume. We observed that even a constant $vc$ with a decelerating target approach profile will produce satisfactory results.

After computing the desired angular and tangential speeds, the control signal is generated as (17):

$$U = Q(\begin{bmatrix} vc \\ \omega c \end{bmatrix}) = G^{-1}(\begin{bmatrix} vc \\ \omega c \end{bmatrix}). \quad (17)$$

Figure-21 shows the overall structure for converting guidance to control.

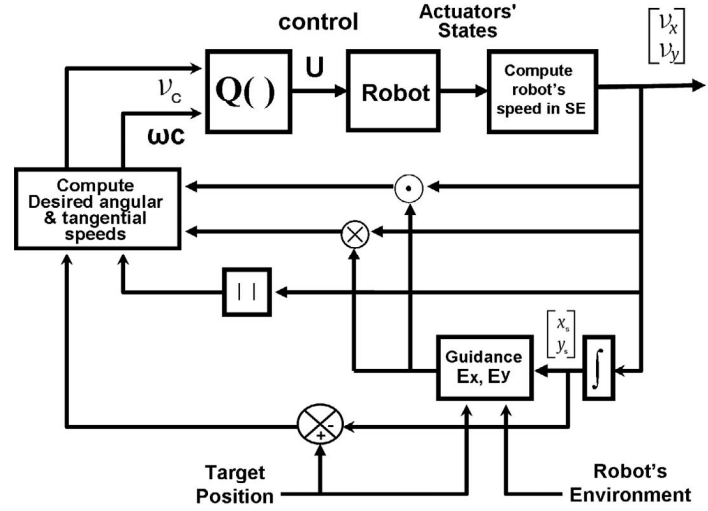

Figure-21: The suggested, hardware-friendly controller

### D. The TCM mobility workflow

This section suggests a structure for TCM that is describe by the flowchart in Figure-22 for integrating the mobility modules in one system. The structure is self-contained where the only piece of information it needs from the external user is where the target is located in the local coordinates of the robot. It provides the robot with full autonomous capabilities without any restrictive assumptions on the robot's space.

Initialization of the structure starts by specifying the coordinates one may use in indexing the representation. The perimeter of the map is a square domain of width D. A Cartesian coordinate system (x,y) is assumed so that the origin is at the center of the domain. Initially, the robot is assumed to lie at the origin (x=0,y=0) of the local coordinates. The positive x-axis is aligned with the principal axis of the robot. ($\theta(0)=0$). The target of the robot is supplied to the mobility structure in the local coordinates of the robot.

The data intake of the structure is from one ultrasonic sensor that is placed in front of the robot and aligned along its principal axis. The continuous output from the sensor (S(t)) provides a measurement of the distance between the sensor and the closest obstacle that lies along the principal axis of the robot. A maximum reading (2.55 meter for the X80) is an indicator that either no obstacle exists along the principal axis or the obstacle is out of sensor range.

The map is recorded on a uniform rectangular grid with a resolution Δ and is stored in an NxN matrix DSE(i,j) (Δ=D/N). If and entry in DSE(i,j) is marked by 1, the location indexed by i and j is considered unsafe. If it is marked by 0, the location is considered possibly safe. DSE is initialized as (18):

$$DSE(1,k) = DSE(N,k) = DSE(k,1) = DSE(k,N) = 1 \quad k = 1,..N, \tag{18}$$
$$DSE(i,j) = 0 \quad i = 2,..N-1, \; j = 2,..N-1$$

At a certain instant in time, given a robot's pose (x,y,θ), DSE is populated as follows (19):

$$Io = \left[ \frac{x + (S + W/2) \cdot \cos(\theta + \delta)}{\Delta} \right], \quad Jo = \left[ \frac{y + (S + W/2) \cdot \sin(\theta + \delta)}{\Delta} \right] \tag{19}$$
$$DSE(Io+m, Jo+n)=1 \quad n=-Im,..Im, \; m=Im,..Im$$
$$N>Io+m>1, \; N>Jo+n>1$$

where Im is a non-negative integer used as a safety margin surrounding the sensed obstacle location, [X] is the rounding integer function of the real number X, 1>>δ>0 and m,n are positive integers used to specify a safety zone around the point (Io,Jo).

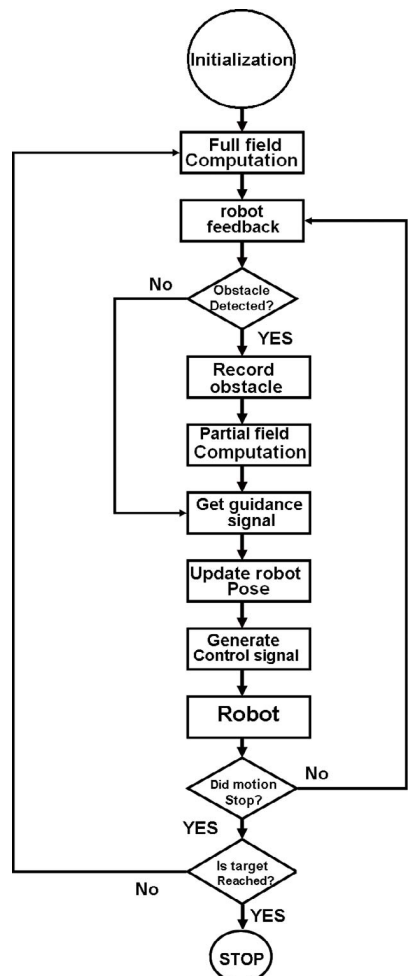

Figure-22: The ultrasonic navigation control structure

Given the initial information that is available, the guidance field is globally computed. If no obstacles are within the sensor range of the robot, the pre-computed guidance information is used. If a new obstacle facing the robot is detected, it is mapped using the procedure in (19) and the guidance field is recomputed around the detected component. The pose of the robot in the local coordinates is updated using its wheels' speeds. This information is combined with the guidance signal to compute the control signals. Since the procedure for constructing the navigation control is provably-correct, the robot should not stop until the target is reached. If the robot stops short of reaching its target, then the cause of the problem has to be the field computation stage. Since there is no motion, computing the guidance field in real-time is no longer important. Therefore, the mobility structure invokes the full field computation stage to correct this problem.

## V. Implementation and Experiments

This section demonstrates the experimental feasibility of the TCM structure suggested in the previous sections. The test emphasizes affordable commercial hardware, inexpensive sensing and low-end processing.

A. Physical setting

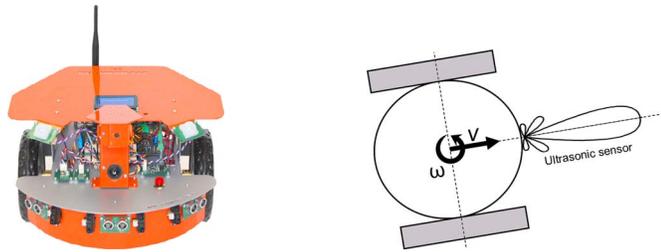

Figure-23: The X80 UGV platform, only the front sensor used in mapping the environment

The platform used is the X80 differential drive robot [81] manufactured by Dr. Robot Inc. Only one of the six ultrasonic sensors of the robot, the front sensor, is used in recording the environment (figure-23). The main lobe of the sensor coincides with the principal axis of the robot. Readings from the sensor are used directly without processing. Cable drums are used in constructing the environment (figure-24). The curved surface of these units causes significant ultrasonic signal distortion.

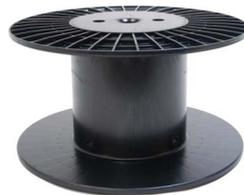

Figure-24: Test environment, partially built from cable drums

The mobility system resides onboard a remote Pentium I5 host running MS Windows 7 (figure-25). The exchange of hard-time sensory and actuation data between the platform and the host is carried-out via wireless IP at a 7 Hz update rate. C# is used for programming. Wireless signals in a restricted cluttered environment suffer from fast fading (figure-26) and causes random momentary interruption of the wireless communication link. Moreover, MS Windows 7 is not a real-time operating system and is highly likely to cause random delays in the servo-link.

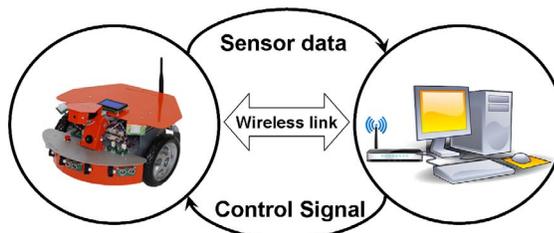

Figure-25: Sensor and actuator signal exchange between X80 and remote host via a wireless link

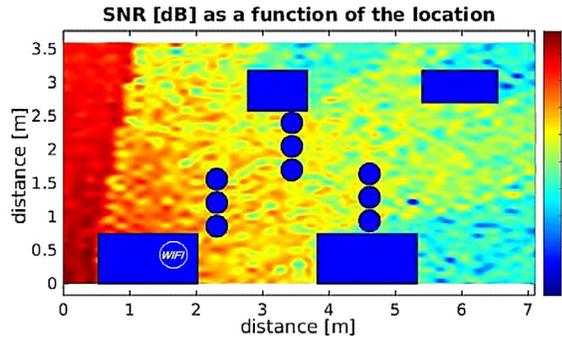
Figure-26: Fast fading in constricted cluttered environment can randomly cause wireless signal outage

Multi-resolution grids [82] can accurately record an environment and maintain an acceptable grid size. However, a simple evenly spaced grid is used to record the sensory data. Advanced deadreckoning techniques [83], even precise optical deadreckoning [84], do exist. However, here, the system uses basic deadreckoning by directly computing the robot's pose from its wheels' speeds (20) with no processing or filtering (21):

$$\begin{bmatrix} v \\ \omega \end{bmatrix} = \begin{bmatrix} \dfrac{r}{2} & \dfrac{r}{2} \\ \dfrac{r}{W} & \dfrac{-r}{W} \end{bmatrix} \begin{bmatrix} \omega_R \\ \omega_L \end{bmatrix}, \quad (20)$$

$$vx = v \cdot \cos(\theta), \quad vy = \sin(\theta) \quad (21)$$
$$x(t+\Delta T) = x(t) + \Delta T \cdot vx(t),$$
$$y(t+\Delta T) = y(t) + \Delta T \cdot vy(t)$$
$$\theta(t+\Delta T) = \theta(t) + \Delta T \cdot \omega(t) \quad x(0)=0, y(0)=0, \theta(0)=0$$

Figure-27 describes the causality of the mobility system that is used in the experiments. This stringent causality forces the harmonic planner to be the source of any action the robot takes. The causality does not allow use of fast reactive corrective action derived from any other module (e.g. the sensing stage). In effect, such usage forces the robot to operate in a cognitive mode all the time.

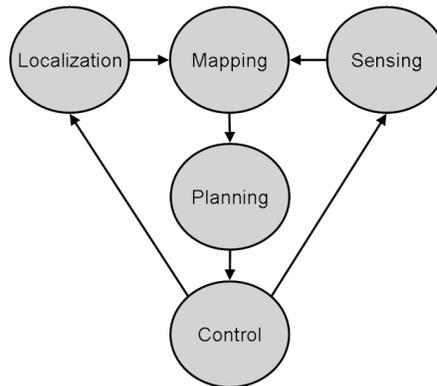
Figure-27: Adopted test causality of the mobility system

The above setting for experimentally testing the suggested structure can seriously disrupt the ability of the mobility system to register and process the environment data. It can also undermine the ability of the robot to take the right action and even cripple this ability altogether when fading causes an outage in the wireless link. It is to be noticed that these conditions are imposed for test purpose only. Better hardware, sensing, filtering and on-board processing may be used if the situation permits.

B. Experimental results

The suggested structure was thoroughly tested by simulation. However, only experimental results based on the X80 platform are reported. Emphasis is placed on sensor-based results under zero *a priori* information about the environment. Selected tests demonstrate the characteristics and capabilities of the mobility structure.

1. Effect of false positive obstacles

This example demonstrates the effect of false detection of environment components on both the ability of the robot to reach its target and the quality of its trajectory. Spurious reflections from raw ultrasonic sensing have the potential to cause erratic behavior. There is also the possibility that an accumulation of these reflections in the representation may cause the belief that the target is unreachable. Figure-28 shows snapshots along with the trajectory of the robot moving to its target. The robot is initially facing the target and has a direct, unobstructed path to its destination. The robot moves directly along a straight line to the target. Figure-29 shows the final guidance field along with the safety events that the sensor discovered. As can be seen, the sensors did not register any safety event during the trip to the target.

Wide turns by the robot can significantly increase false registrations by the ultrasonic sensors. Figure-30 shows snapshots and the trajectory the robot takes to the target. The robot is in the same position as in figure-28 with the exception that its heading makes a 90-degree angle with respect to the target. The robot's initial heading is 90 degree out of phase with the guidance vector. This forces the robot to make a large turn in order to synchronize with guidance. Figure-31 shows the final guidance field superimposed on sensor registration. One can observe that false registrations blocks 50% of the passage to the target. Despite this large error in recording the environment, the robot still moves to the target along a well-behaved trajectory experiencing only minor deviation from the direct trajectory in figure-28.

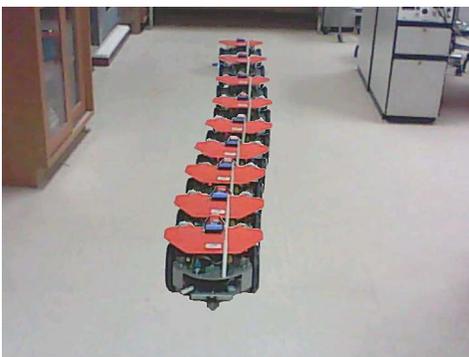 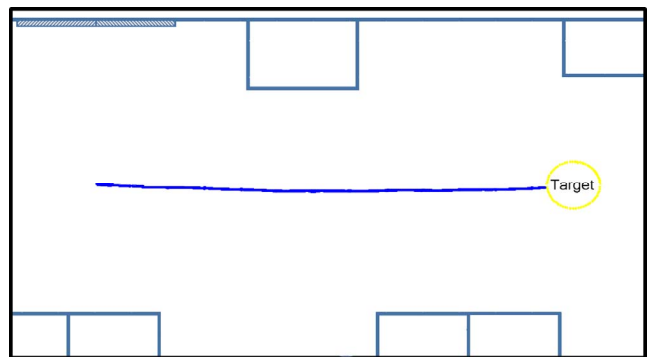

Figure-28: Trajectory to the target, obstruction-free, initial angle = 0

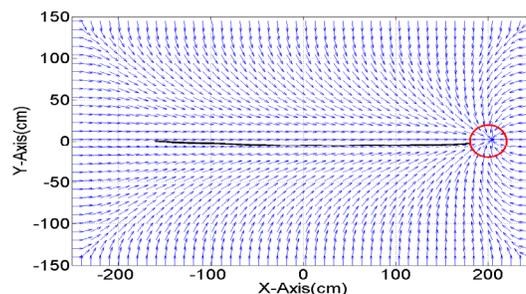

Figure-29: Final guidance policy and environment registration corresponding to figure-28

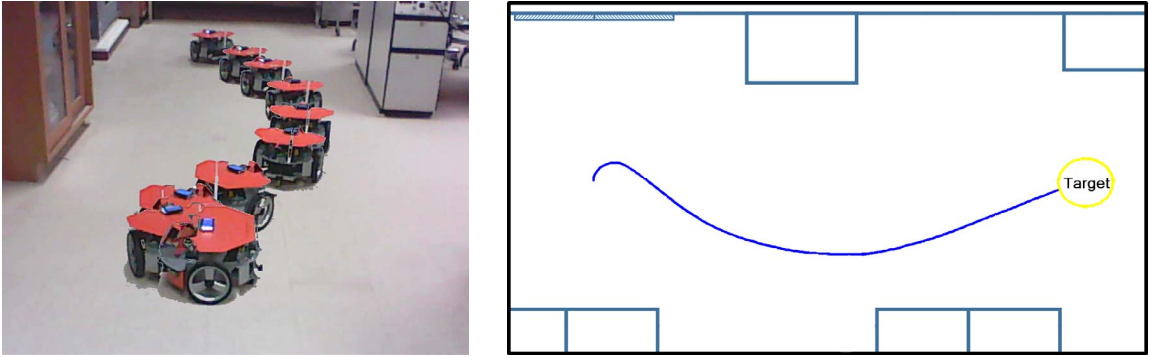
Figure-30: Trajectory to the target, obstruction-free, initial angle = 90

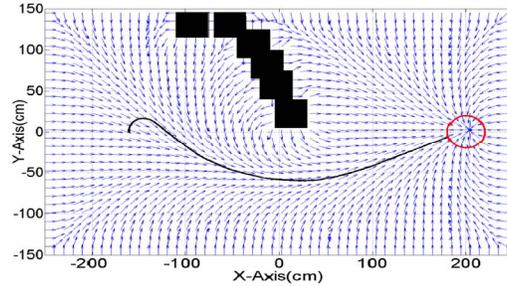
Figure-31: Final guidance policy and environment registration corresponding to figure-30

Figure-32 shows the ultrasonic sensor activities versus time. Figure-33 shows the linear velocity and the angular velocity of the robot respectively. Notice that after the short transient period needed for the robot to synchronize with the guidance signal, the linear velocity remains around a constant value. The mobility system also maintains the angular velocity close to zero. Figure-34 shows the control velocity signal of the right and left wheels of the robot. The control signals are steady and well-behaved.

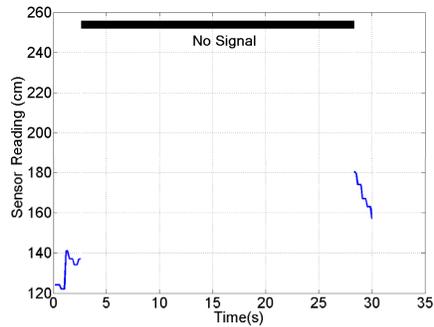
Figure-32: Sensor activities corresponding to trajectory in figure-30

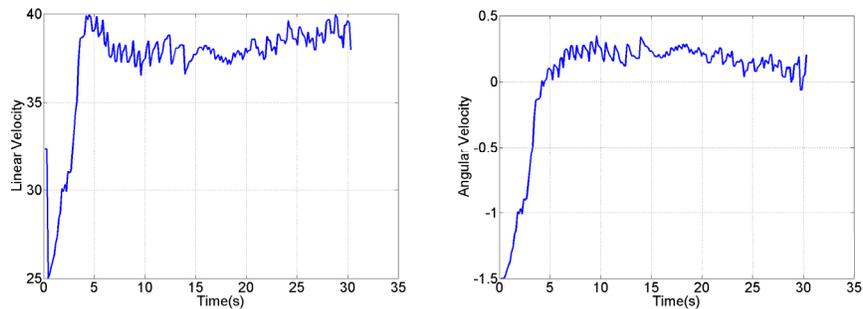
Figure-33: Tangential and angular velocities corresponding to the trajectory in figure-30

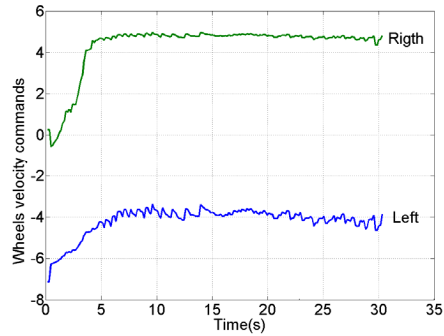
Figure-34: Actuation signals corresponding to the trajectory in figure-30

2. Obstacle course: sensor-based

An obstacle course made of cable drums is placed between the robot and the target. The obstacles force the robot to change both curvature and orientation. The robot has zero *a priori* information about the environment. Figure-35 shows snapshots of the robot along with the corresponding trajectory. The trajectory is well-behaved and moves directly to the target except for a minor detour at the start of motion. Figure-36 shows the final guidance field with the sensed hazardous zone. Although the belief safety map significantly differs from the physical environment, the resulting trajectory accommodates well the geometry of the environment. Figure-37 shows snapshots of the guidance fields at different instants in time with the hazard zones superimposed on it. One can easily notice the incremental and noncommittal nature the HPF-based guidance policy.

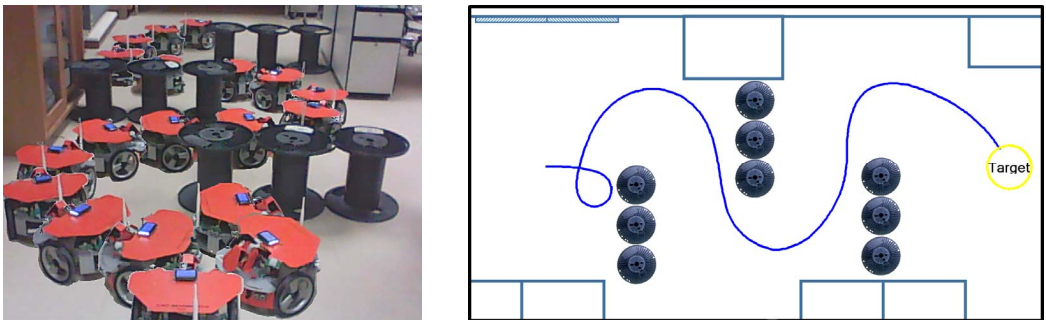
Figure-35: Trajectory to the target, obstacle course, sensor-based

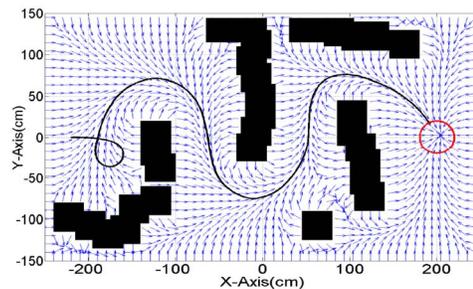
Figure-36: Final guidance policy and environment registration corresponding to figure-35

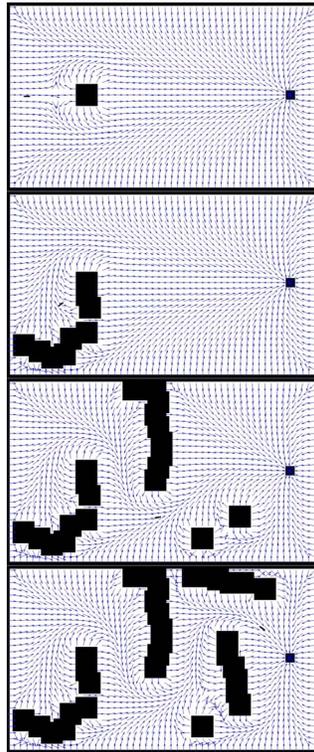

Figure-37: Snapshots of the guidance policy and environment registration corresponding to figure-35

Figure-38 shows the signal from the ultrasonic sensor, which is discontinuous and unavailable more than 50% of the duration of the mission. Figure-39 shows the tangential and angular velocities of the robot. The behavior of these velocities closely follows the pattern in the previous cases. After a short transient period, the control signal synchronized the robot with the guidance signal. The tangential velocity of the robot converges to a narrow band around a constant value and the angular velocity stays around zero. The control signals are in figure-40. As can be seen, the right and left velocities of the robot are well-behaved.

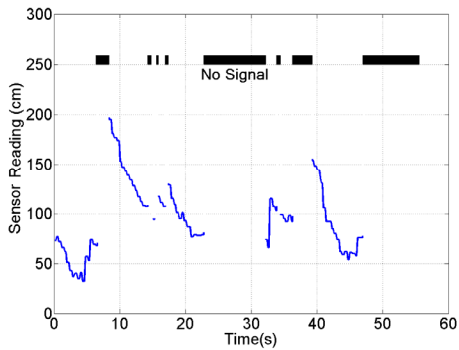

Figure-38: Sensor activities corresponding to trajectory in figure-35

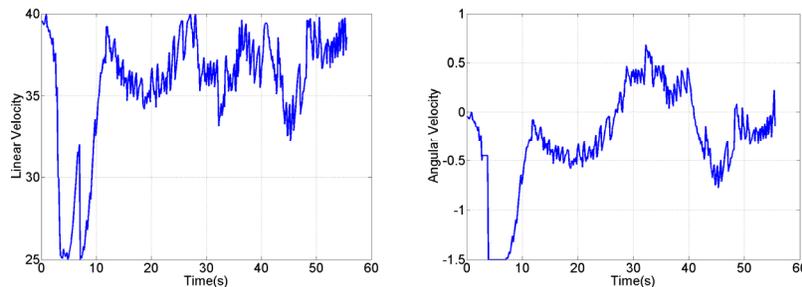

Figure-39: Tangential and angular velocities corresponding to the trajectory in figure-35

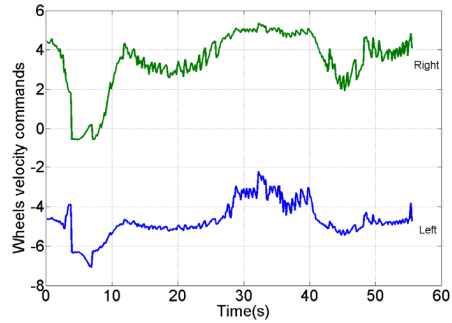
Figure-40: Actuation signals corresponding to the trajectory in figure-35

3. Obstacle course: Model-based

Although the mobility system can function even if the environment is totally unknown, it has the ability to accommodate *a priori* information if it is available to enhance performance. The obstacle course example is repeated with no on-line sensory acquisition. Here, the initial belief-based representation contains some components of the environment. Figure-41 shows snapshots of the robot and the corresponding trajectory. Figure-42 shows the trajectory superimposed on the guidance field and the *a priori* information contained in the representation. As can be seen, the trajectory is well-behaved, keeps good clearance from the obstacles and is almost optimal.

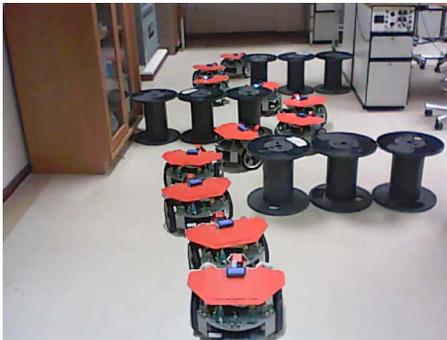 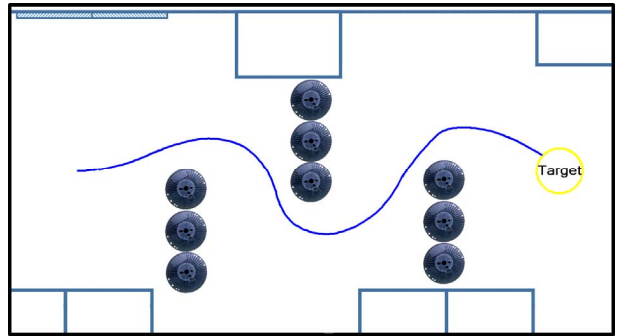
Figure-41: Trajectory to the target, obstacle course, model-based

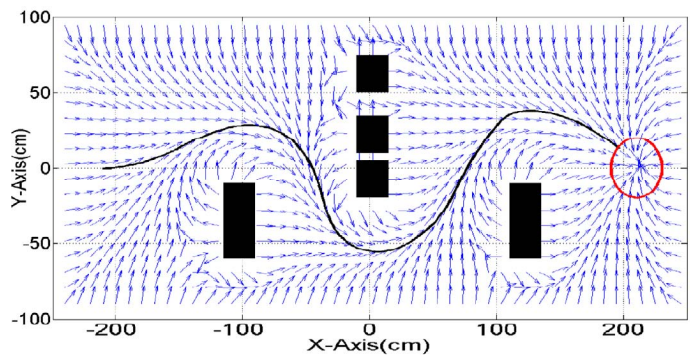
Figure-42: Guidance policy and environment registration corresponding to figure-41

Figure-43 shows the tangential and angular velocities of the robot; previous observations hold. Figure-44 shows the actuation signals. Well-behaved actuation signals resulted with a dynamic range less than that of their sensor-based counterpart. More importantly, the use of partial *a priori* information did not only improve the quality of the trajectory and reduced stress on the actuator, it also considerably speeded-up mission execution. In the sensor–based case, it took the robot almost 60 seconds to reach the target compared to 30 seconds for the case utilizing partial a *priori* information.

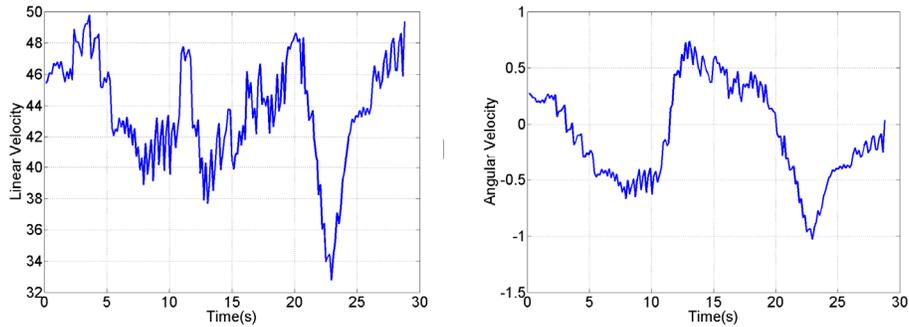
Figure-43: Tangential and angular velocities corresponding to the trajectory in figure-41

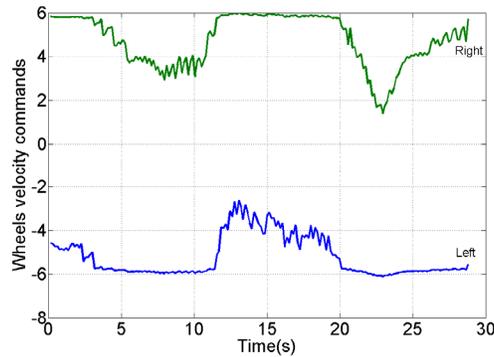
Figure-44: Actuation signals corresponding to the trajectory in figure-41

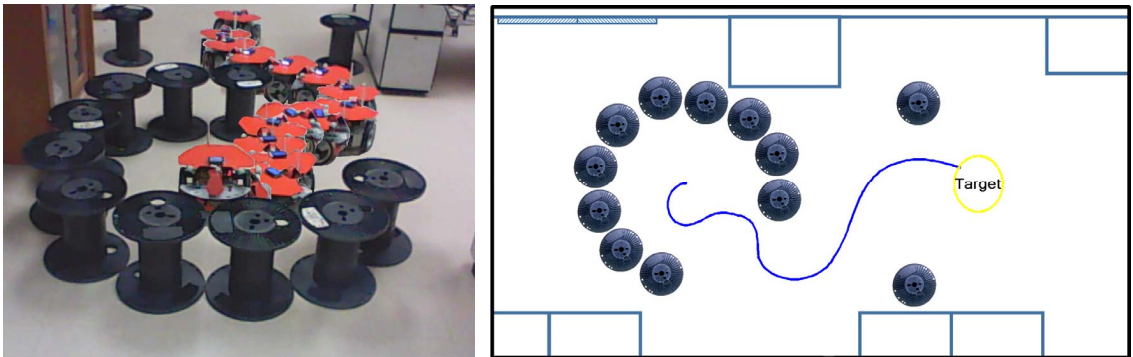
Figure-45: Trajectory to the target, trap, angle=180

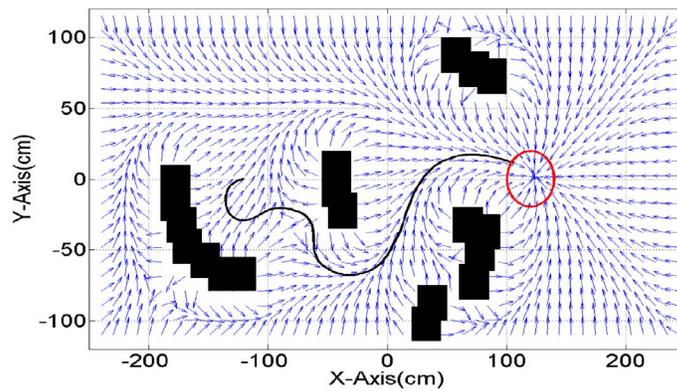
Figure-46: Final guidance policy and environment registration corresponding to figure-45

4. Traps

The guidance and control modules are provably-correct. Theoretically speaking, the mobility system should be able to navigate any environment regardless of its geometry or topology. If the target is potentially accessible and the robot is not able to reach it, an implementation issue is the one responsible for the failure. Many researchers consider trap situations as a strong test of the capabilities of a mobility system. In this example, an explicit trap is constructed and the robot is initially oriented in an opposite direction to the target, 180º out of phase with guidance. Figure-45 shows snapshots of the robot and the trajectory along which it moved to the target. Figure-46 shows the trajectory superimposed on the final guidance field and the discovered environments components. Figure-47 shows snapshots of the guidance field at different instants in time. The signal from the ultrasonic sensor is in figure-48, the tangential and angular velocities of the robot are in figure-49 and the actuation signals are in figure-50. As can be seen, the robot successfully dealt with this situation. The trajectory, differential properties and control signals behaved in a manner similar to the previous environment settings. Different robot orientations (Figure-51, 52) yielded results similar to those in figure-45.

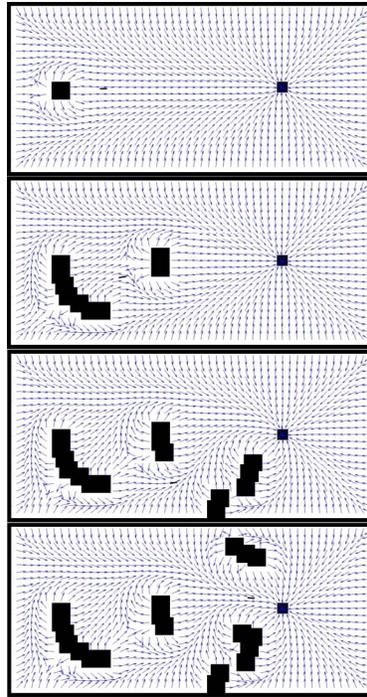

Figure-47: Snapshots of the guidance policy and environment registration corresponding to figure-45

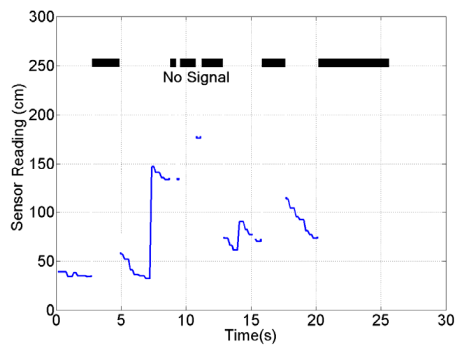

Figure-48: Sensor activities corresponding to trajectory in figure-45

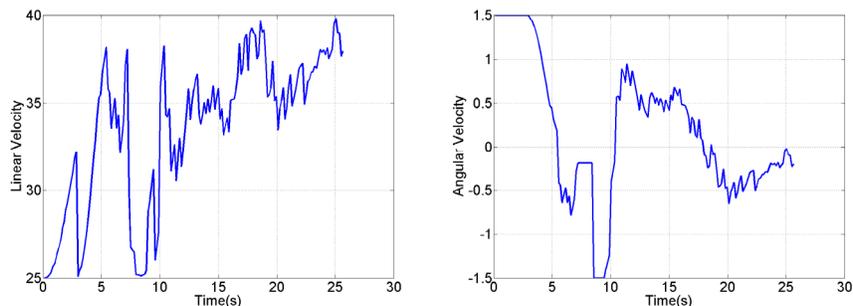

Figure-49: Tangential and angular velocities corresponding to the trajectory in figure-45

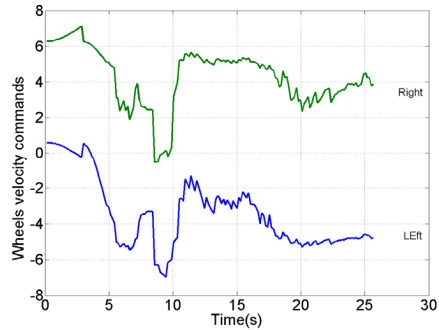
Figure-50: Actuation signals corresponding to the trajectory in figure-45

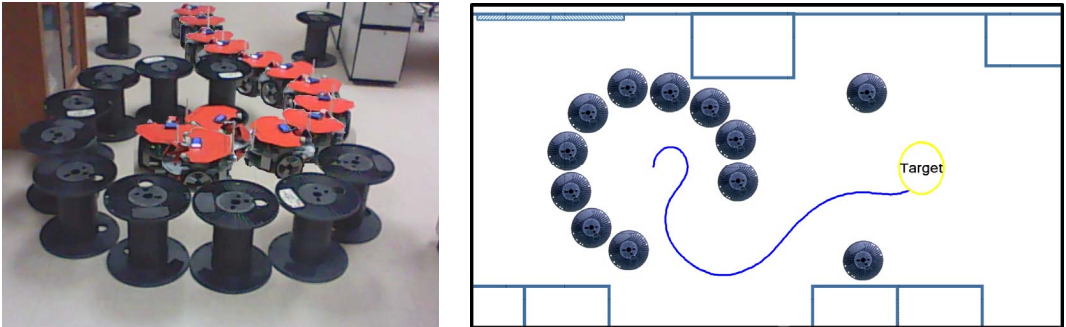
Figure-51: Trajectory to the target, trap, angle=90

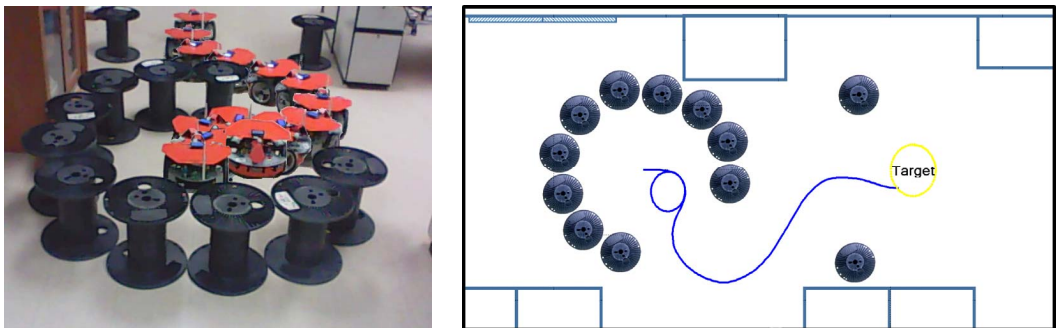
Figure-52: Trajectory to the target, trap, angle=0

5. Modulated vs. constant tangential velocities

The mobility system distributes safety on its modules where safety at the system level is an emergent state. Safety is introduced in the actuation module by requiring the control signal to modulate the tangential velocity of the robot based on it relative heading with respect to the guidance signal. The mobility system is tested for the case when guidance is converted directly to control without modulating the tangential speed. The results with no modulation are satisfactory. However, the robot is more hyperactive compared to the modulated case. This example demonstrates the positive impact velocity modulation has on mobility as a whole. A simple environment created from one cable drum is used in the comparison between the modulated and un-modulated cases. Figure-53 and figure-54 show robot snapshots and corresponding trajectories for the unmodulated and modulated cases respectively. Figure-55 shows the sensor activities and figure-56 shows the control signals for the unmodualted and modulated cases respectively. The trajectory produced by modulating the tangential velocity has less fluctuation and keeps better distance away from the obstacles. One can also notice that the modulated case has less sensory activities and the dynamic range of the control signals is lower for the modulated case compared to the unmodualted one. The trip time for the modulated case (25 seconds) is less than that for the unmodulated case (35 seconds).

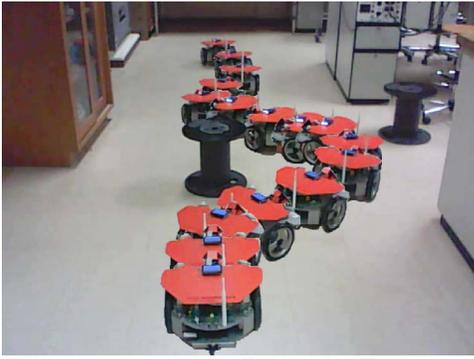 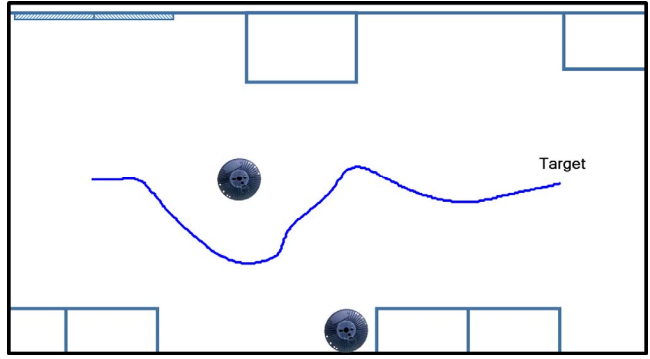

Figure-53: Trajectory to the target, constant tangential velocity

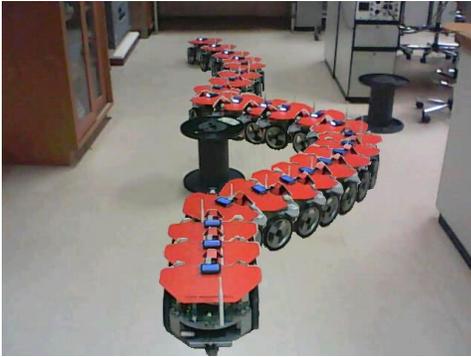 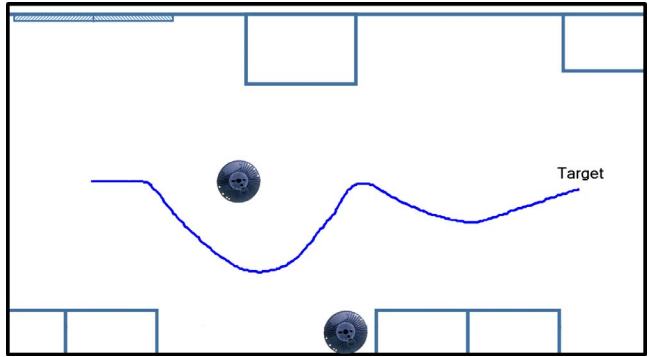

Figure-54: Trajectory to the target, modulated tangential velocity

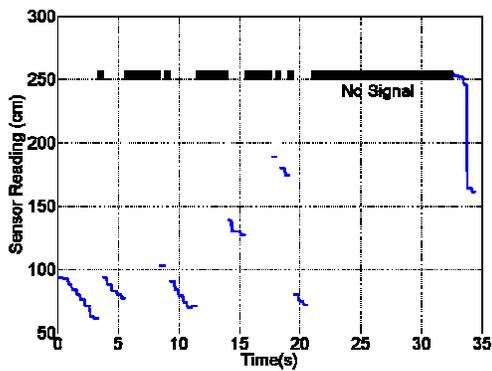 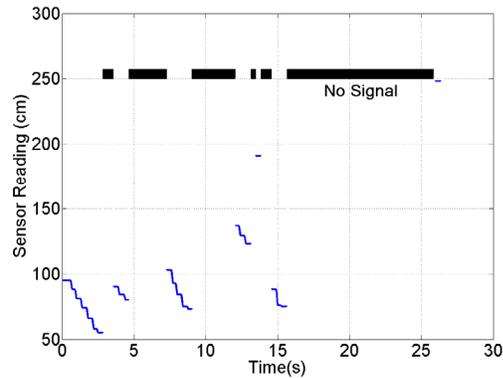

Figure-55: Sensor activities, constant (figure-53) and modulated velocity (figure-54) cases respectively

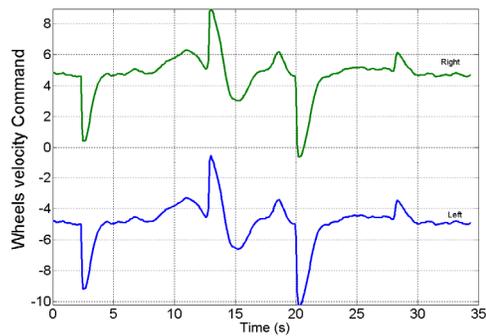 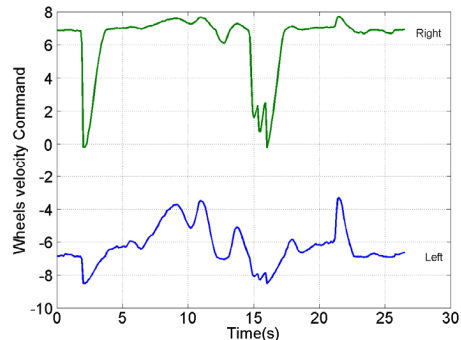

Figure-56: Actuation signals, constant (figure-53) and modulated velocity (figure-54) cases respectively

## VI. Conclusions

This paper provides a proof of principle that TCM is possible using affordable hardware, modest sensing and does not require a dedicated exploration and mapping stage. It also demonstrates that the harmonic potential field approach to planning is suitable for sensor-based motion planning at the servo-level of a mobile robot. The work in this paper provides a strong reason to re-examine the belief that accurate, spatial mapping is a prerequisite for navigating satisfactorily an unstructured environment. A properly designed mobility system can provide good performance, even if the map is incomplete and inaccurate. The work also provides a good reason to consider seriously the issue of joint design of mobility modules as a means of obtaining reliable system performance at a reasonable hardware cost.

ACKNOWLEDGEMENT

The authors gratefully acknowledge the assistance of King Fahad University of Petroleum and Minerals. The authors also acknowledge the assistance of Mr. Mohanad Ahmed with the implementation of the experiments. The authors are also grateful to Mr. Michael Joseph Fogarty for the excellent job he did in proofreading the manuscript.